\documentclass[12pt]{spieman}  
\usepackage{amsmath,amsfonts,amssymb}
\usepackage{graphicx}
\usepackage{setspace}
\usepackage{tocloft}
\title{Simulation-Driven Deep Learning Framework for Raman Spectral Denoising Under Fluorescence-Dominant Conditions}

\author[a]{Mengkun Chen}
\author[a]{Sanidhya D. Tripathi}
\author[a,*]{James W. Tunnell}
\affil[a]{The University of Texas at Austin, Department of Biomedical Engineering, 107 W Dean Keeton St., Austin, Texas, US}

\cftpagenumbersoff{figure}
\cftpagenumbersoff{table}
\begin{document} 
\maketitle

\begin{abstract}
Raman spectroscopy enables non-destructive, label-free molecular analysis with high specificity, making it a powerful tool for biomedical diagnostics. However, its application to biological tissues is challenged by inherently weak Raman scattering and strong fluorescence background, which significantly degrade signal quality. In this study, we present a simulation-driven denoising framework that combines a statistically grounded noise model with deep learning to enhance Raman spectra acquired under fluorescence-dominated conditions. We comprehensively modeled major noise sources. Based on this model, we generated biologically realistic Raman spectra and used them to train a cascaded deep neural network designed to jointly suppress stochastic detector noise and fluorescence baseline interference. To evaluate the performance of our approach, we simulated human skin spectra derived from real experimental data as a validation case study. Our results demonstrate the potential of physics-informed learning to improve spectral quality and enable faster, more accurate Raman-based tissue analysis.
\end{abstract}

\keywords{Raman spectroscopy, denoising, deep learning, noise modeling}

{\noindent \footnotesize\textbf{*}James W. Tunnell,  \linkable{jtunnell@mail.utexas.edu} }

\begin{spacing}{2}   

\section{Introduction}
\label{sect:intro}  
Raman spectroscopy is highly specific, allowing for the non-destructive analysis of biological samples without the need for labels or dyes \cite{Smith2005}. Its ability to provide detailed molecular information has made it invaluable in fields such as chemistry, materials science, and biomedical imaging.

In biomedical applications, Raman scattering has emerged as a critical technique for diagnosing diseases, including cancer, by detecting molecular changes associated with pathological states \cite{Kneipp1999, Wang2017}. The high sensitivity of Raman spectroscopy to biochemical alterations enables the identification of disease biomarkers, making it a promising tool for early detection and monitoring of conditions such as skin cancer. \cite{Xu_biophysical, Chen2022_deep, Zhang2020_clinical}

Despite its high specificity and non-destructive nature, Raman scattering has notable limitations. The primary drawback is its inherently weak signal, as only a small fraction of incident photons undergo Raman scattering\cite{Long2002, Smith2005}. This weak scattering results in a low signal-to-noise ratio (SNR), which typically necessitates long integration times to acquire spectra of acceptable quality. The challenge is especially pronounced in biological tissues, where strong fluorescence background—arising from endogenous fluorophores—often dominates the collected signal \cite{Zeng1999, Alfano1997, Kong2015}. This intense autofluorescence introduces broad baseline distortions and photon shot noise, which further degrades SNR, effectively masking the already weak Raman features. 

In Raman spectroscopy, the overall noise affecting signal quality can be broadly categorized into stochastic noise and baseline background interference. Stochastic noise includes photon shot noise, which arises from the probabilistic nature of photon arrival \cite{Bocklitz2016}; dark (thermal) noise, generated by thermally excited electrons in the absence of illumination; and read noise, introduced during the detector’s charge-to-voltage conversion and digitization process \cite{Smith2005, Zhao2007}. Additionally, fixed pattern noise—due to pixel-to-pixel response variation—can introduce consistent spatial artifacts over repeated measurements \cite{Sarder2006}. Baseline background interference is primarily caused by fluorescence, which is particularly dominant in biological tissues due to the presence of endogenous fluorophores \cite{Alfano1997, Kong2015}. Although stray light from Rayleigh scattering or ambient sources may also contribute to the baseline, it is typically well suppressed in biomedical Raman systems through appropriate filtering and controlled environments \cite{Matousek2005, Keller2010}, and can therefore be reasonably neglected in simulations of biological samples. Thus, the central task in Raman denoising is to effectively suppress both the stochastic detector noise and the baseline background interference to recover clean and interpretable spectra.

Various denoising strategies have been developed to address the noise and background challenges in Raman spectroscopy. Traditional approaches, such as polynomial fitting and baseline correction methods, have been extensively used.~\cite{Lieber2003_FluorSub, SavitzkyGolay1964_SG, Eilers2005_AsLS, Zhang2010_airPLS, Baek2015_arPLS, Han2024_Review} However, these techniques often fail to handle complex fluorescence backgrounds effectively~\cite{jahn2021noise}.
    
Spectral filtering methods, such as the Savitzky-Golay (SG) filter and wavelet filter, have been commonly used for noise reduction, as they preserve peak shapes while suppressing high-frequency noise.~\cite{Barton2018_OptimalDenoisingRaman, Chen2014_WienerRaman, Han2024_Review} More advanced approaches, including wavelet transform techniques, have enabled selective removal of noise by decomposing signals into frequency components~\cite{lin2021microsecond}. Yet, such methods may struggle in scenarios where noise and signal components overlap significantly.

Machine learning and adaptive decomposition methods have become increasingly prominent for Raman denoising. For instance, recent studies have demonstrated the effectiveness of physics-based noise estimation methods combined with machine learning, such as Empirical Mode Decomposition (EMD) and Variational Mode Decomposition (VMD), in addressing non-stationary noise~\cite{dragomiretskiy2014variational}.

Deep learning techniques have emerged as transformative tools in Raman spectroscopy. They have significantly improved SNR and resolution in hyperspectral Raman imaging and nanoscopy while enabling fast, low-laser-power measurements that minimize sample damage~\cite{chen2021deep, lin2021microsecond}. Moreover, noise learning techniques using deep learning architectures, such as attention U-Net models, have been shown to enhance SNR by up to tenfold and reduce mean-square error by nearly 150-fold. These noise learning approaches leverage physics-based data generation, allowing them to bypass the need for large labeled datasets and adapt directly to instrument-specific noise patterns~\cite{he2024noise}. These methods have shifted the focus from sample-dependent learning to instrument-dependent learning, marking a significant step forward in the robustness and versatility of Raman denoising.

While deep learning has shown significant promise for Raman denoising, its success depends heavily on the availability of clean spectra for supervision. A common strategy to obtain ground truth spectra involves using long integration times to suppress stochastic noise \cite{lin2021microsecond, Fu2020Stochastic}. However, this approach is inherently time-consuming and may not be practical in clinical or dynamic experimental settings \cite{Kirsch2022ClinicalRaman, Lu2017ClinicalSpeed}. Moreover, it is highly sample-dependent, making it difficult to generalize across different biological contexts or measurement systems \cite{he2024noise, Shao2020AdditiveNoise}.

To circumvent the need for prolonged acquisitions, some studies have proposed generating synthetic training data by adding noise to clean Raman signals \cite{Shao2020AdditiveNoise, Liu2018AdditiveSim}. In this approach, clean spectra—often obtained from literature or idealized models—are superimposed with estimated noise patterns. Yet, this additive model of noise is fundamentally limited. In real detectors, noise characteristics are signal-dependent and follow complex distributions such as Poisson or Gaussian mixtures, depending on photon statistics, sensor gain, and background levels \cite{Zhao2007SensorNoise}. Simply adding sampled noise to clean spectra ignores these interactions and may lead to unrealistic data distributions \cite{dragomiretskiy2014variational}.

To improve upon purely additive noise models, some researchers have attempted to extract noise components from experimental data using unsupervised decomposition techniques. For example, He et al. (2024) employed singular value decomposition (SVD) to isolate noise components from raw Raman measurements, which were then used to train denoising networks \cite{he2024noise}. While this approach can provide a more data-driven noise estimate, it introduces subjectivity in choosing which components represent noise versus signal. The thresholding and component selection in SVD are not rigorously defined, making it difficult to generalize across datasets or instruments. As a result, such decompositions may still produce biased or unrealistic noise profiles, especially when the signal-to-noise ratio is low or the signal overlaps in structure with the noise.

Moreover, most existing denoising studies primarily focus on reducing stochastic noise while largely neglecting the fluorescence background, which is often the dominant interference in biological samples. In real-world tissue spectroscopy, fluorescence can exceed the Raman signal by orders of magnitude and manifest as a broad, nonlinear baseline that is difficult to separate from the true spectral features \cite{Zeng1999, Alfano1997, Kong2015}. Many deep learning-based and traditional denoising methods assume relatively low background conditions or preprocessed data, thereby limiting their applicability to in vivo or clinical scenarios \cite{Jiang2022DeepRaman, Fu2020Stochastic}. Without explicitly modeling or removing the fluorescence baseline, these methods may leave behind residual background artifacts that impair downstream analysis and interpretation.

In this work, we propose a simulation-driven denoising framework that integrates a statistically grounded noise model with deep learning to reconstruct Raman spectra under biologically realistic conditions. We built a comprehensive statistical model to characterize the noise sources in Raman spectroscopy, including photon shot noise, dark noise, read noise, sensor gain, sensor offset and fluorescence interference. Using this model, we simulated biologically relevant Raman spectra and fluorescence backgrounds. A cascaded deep learning architecture was then employed to jointly learn stochastic noise patterns and suppress the dominant fluorescence background, resulting in more accurate and robust spectral reconstruction. Finally, we used real-spectrum-based simulated human skin spectra as a case study to validate the effectiveness of our denoising process.

\section{Materials and Methods}

In this section, we detail the construction of our noise model, the simulation of Raman spectra and fluorescence interference, the procedure for collecting real noise data, and the process for generating simulated human skin spectra.

\subsection{Spectral Signal Model with Noise}
\label{sect:noise model}

The signal collected by the sensor can be modeled as:

\begin{equation}
    Y(\lambda) = g(\lambda) \cdot 
    \underbrace{[R(\lambda) + F(\lambda) + D(\lambda)]}_{\mu(\lambda)} + Rd(\lambda) + O(\lambda)
\end{equation}

Here, \( Y(\lambda) \) is the raw signal collected by the camera sensor at spectral wavenumber \( \lambda \) (in \( \text{cm}^{-1} \)). \( g(\lambda) \) is the sensor gain, and \( \mu(\lambda) \) is the total number of photoelectrons generated from three sources: \( R(\lambda) \) for Raman-scattered photons, \( F(\lambda) \) for fluorescence photons, and \( D(\lambda) \) for thermally generated dark noise electrons. \( Rd(\lambda) \) is the read noise, and \( O(\lambda) \) is the fixed-pattern offset from the sensor.

We model the photon/electron generation process as a Poisson process:

\begin{equation}
    \mu(\lambda) \sim \text{Poisson}(\hat{\mu}(\lambda))
\end{equation}

where \( \hat{\mu}(\lambda) \) is the expected photon/electron count at wavenumber \( \lambda \). For sufficiently large values of \( \hat{\mu}(\lambda) \) (e.g., \( \hat{\mu}(\lambda) \geq 30 \)), the Poisson distribution can be approximated by a Gaussian:

\begin{equation}
    \mu(\lambda) \sim \text{Poisson}(\hat{\mu}(\lambda)) \approx \mathcal{N}(\hat{\mu}(\lambda), \hat{\mu}(\lambda))
\end{equation}

After applying the sensor gain \( g(\lambda) \), the distribution becomes:

\begin{equation}
    g(\lambda) \cdot \mu(\lambda) \sim \mathcal{N}(g(\lambda) \cdot \hat{\mu}(\lambda),\ g^2(\lambda) \cdot \hat{\mu}(\lambda))
\end{equation}

The read noise is modeled as a zero-mean Gaussian:

\begin{equation}
    Rd(\lambda) \sim \mathcal{N}(0, \sigma_{\text{read}}^2(\lambda))
\end{equation}

The sensor offset is modeled as a Gaussian with mean \( \hat{O}(\lambda) \) and variance \( \sigma_O^2(\lambda) \):

\begin{equation}
    O(\lambda) \sim \mathcal{N}(\hat{O}(\lambda), \sigma_O^2(\lambda))
\end{equation}

Thus, the total observed signal \( Y(\lambda) \) follows:

\begin{equation}
    Y(\lambda) \sim \mathcal{N}\left(g(\lambda) \cdot \hat{\mu}(\lambda) + \hat{O}(\lambda),\ g^2(\lambda) \cdot \hat{\mu}(\lambda) + \sigma_{\text{read}}^2(\lambda) + \sigma_O^2(\lambda)\right)
\end{equation}

Expanding \( \hat{\mu}(\lambda) = \hat{R}(\lambda) + \hat{F}(\lambda) + \hat{D}(\lambda) \), we write:

\begin{align}
\label{eq: whole spectra model}
    Y(\lambda) \sim \mathcal{N}\Big(&g(\lambda) \cdot \big(\hat{R}(\lambda) + \hat{F}(\lambda) + \hat{D}(\lambda)\big) + \hat{O}(\lambda), \notag \\
    &g^2(\lambda) \cdot \big(\hat{R}(\lambda) + \hat{F}(\lambda) + \hat{D}(\lambda)\big) + \sigma_{\text{read}}^2(\lambda) + \sigma_O^2(\lambda)\Big)
\end{align}

We estimate the sensor gain \( g(\lambda) \) using a system response calibration curve obtained from a reference sample (see Section \ref{sec: system response}). By dividing the measured signal by this gain curve, we normalize the signal:

\begin{align}
\label{eq: whole spectra model after cali naive}
    Y_{\text{cali}}(\lambda) = \frac{Y(\lambda)}{g(\lambda)} 
    \sim \mathcal{N}\Big(&\hat{R}(\lambda) + \hat{F}(\lambda) + \hat{D}(\lambda) + \hat{O}'(\lambda), \notag \\
    &\hat{R}(\lambda) + \hat{F}(\lambda) + \hat{D}(\lambda) + \sigma_{\text{read}}'^2(\lambda) + \sigma_O'^2(\lambda)\Big)
\end{align}

where

\[
\hat{O}'(\lambda) = \frac{\hat{O}(\lambda)}{g(\lambda)}, \quad \sigma_{\text{read}}'^2(\lambda) = \frac{\sigma_{\text{read}}^2(\lambda)}{g^2(\lambda)}, \quad \sigma_O'^2(\lambda) = \frac{\sigma_O^2(\lambda)}{g^2(\lambda)}
\]

To separate fluorescence sources, we decompose:

\begin{align}
\label{eq: whole spectra model after cali}
    Y_{\text{cali}}(\lambda) \sim \mathcal{N}\Big(&\hat{R}(\lambda) + \hat{F}_{\text{sample}}(\lambda) + \hat{F}_{\text{comp}}(\lambda) + \hat{D}(\lambda) + \hat{O}'(\lambda), \notag \\
    &\hat{R}(\lambda) + \hat{F}_{\text{sample}}(\lambda) + \hat{F}_{\text{comp}}(\lambda) + \hat{D}(\lambda) 
    + \sigma_{\text{read}}'^2(\lambda) + \sigma_O'^2(\lambda)\Big)
\end{align}

We then collect a dark frame with the laser on but no sample (Section \ref{sec: dark frame}). The signal from this dark frame is:

\begin{equation}
\label{eq: dark fram after cali}
    Y_{\text{cali}}^{dl}(\lambda) \sim \mathcal{N}\left(\hat{F}_{\text{comp}}(\lambda) + \hat{D}(\lambda) + \hat{O}'(\lambda),\ \hat{F}_{\text{comp}}(\lambda) + \hat{D}(\lambda) + \sigma_{\text{read}}'^2(\lambda) + \sigma_O'^2(\lambda)\right)
\end{equation}

Subtracting this dark frame from the calibrated signal gives:

\begin{align}
\label{eq: final model}
    S(\lambda) = Y_{\text{cali}}(\lambda) - Y_{\text{cali}}^{dl}(\lambda) 
    \sim \mathcal{N}\Big(&\hat{R}(\lambda) + \hat{F}_{\text{sample}}(\lambda), \notag \\
    &\hat{R}(\lambda) + \hat{F}_{\text{sample}}(\lambda) + 
    2 \cdot \big(\hat{F}_{\text{comp}}(\lambda) + \hat{D}(\lambda) 
    + \sigma_{\text{read}}'^2(\lambda) + \sigma_O'^2(\lambda)\big) \Big)
\end{align}

Letting:

\[
S_{\text{sample}}(\lambda) = \hat{R}(\lambda) + \hat{F}_{\text{sample}}(\lambda), \quad
S_{\text{dark}}(\lambda) = \hat{F}_{\text{comp}}(\lambda) + \hat{D}(\lambda) + \sigma_{\text{read}}'^2(\lambda) + \sigma_O'^2(\lambda)
\]

we obtain the final model:

\begin{equation}
\label{eq: final model simple}
    S(\lambda) \sim \mathcal{N}\left(S_{\text{sample}}(\lambda),\ S_{\text{sample}}(\lambda) + 2 \cdot S_{\text{dark}}(\lambda)\right)
\end{equation}

In other words, to simulate a realistic Raman signal with noise, we generate the true Raman signal \( \hat{R}(\lambda) \) and sample-dependent fluorescence \( \hat{F}_{\text{sample}}(\lambda) \), and use the dark frame to estimate \( S_{\text{dark}}(\lambda) \). This allows us to sample \( S(\lambda) \) from the distribution in Eq.~\ref{eq: final model simple}.

\subsection{System Response Calibration}
\label{sec: system response}
To accurately interpret the measured Raman signal, it is essential to correct for the wavelength-dependent sensitivity of the detection system. This sensitivity is influenced by a range of system-specific factors, including the quantum efficiency of the detector, optical transmission losses, and grating efficiency. Together, these factors define the system response function, which modulates the true spectral signal across wavenumbers. Without calibration, the raw measured intensity is not directly proportional to the actual photon flux, which introduces bias into both qualitative and quantitative spectral analysis.

System response calibration allows us to compensate for this wavelength-dependent variation by estimating and removing the effect of the system gain function \( g(\lambda) \). Once the gain function is characterized, the measured signal \( Y(\lambda) \) can be normalized by dividing by \( g(\lambda) \), yielding a response-corrected signal that reflects the actual photon count entering the system. This process is particularly critical in simulation-based modeling, where accurate estimation of photon counts is needed to apply Poisson or Gaussian noise models realistically.

In our work, we use a reference sample with a known and traceable spectral output to characterize the system response. Specifically, we employ a NIST(National Institute of Standards and Technology, Standard Reference Material \#2246)-traceable calibrated light source, which emits a smooth and well-characterized spectrum over the relevant wavenumber range. By measuring the spectrum of this reference under the same experimental conditions, we obtain a measured response \( Y_{\text{ref}}(\lambda) \), which is then compared to the true known spectral radiance \( R_{\text{true}}(\lambda) \). The system gain function is computed as:

\begin{equation}
    g(\lambda) = \frac{Y_{\text{ref}}(\lambda)}{R_{\text{true}}(\lambda)}
\end{equation}

Once \( g(\lambda) \) is known, we apply it to all subsequent measurements via normalization:

\begin{equation}
    Y_{\text{cali}}(\lambda) = \frac{Y(\lambda)}{g(\lambda)}
\end{equation}

This calibration ensures that all simulated and real data are referenced to the same physical scale, making it possible to apply our statistical noise model consistently across the spectrum. Furthermore, it improves the fidelity of downstream signal separation and denoising by reducing systematic intensity artifacts caused by the detection system.

\subsection{Dark Frame Collection}
\label{sec: dark frame}
We collected Raman noise spectra using a custom-built fiber-coupled Raman microscope, as described in detail by Feng \textit{et al.}~\cite{Xu_biophysical}. The system consists of a 830~nm diode laser for excitation, a backscattering collection geometry, and a thermoelectrically cooled CCD detector coupled to a spectrograph for high-sensitivity spectral acquisition. To isolate the intrinsic system and environmental noise characteristics, we acquired noise-only spectra in a dark room with the laser turned on and no sample placed beneath the objective lens. This configuration ensures that the measured signal contains only dark current, read noise, fixed-pattern offset and some intrinsic fluorescence from optical components themselves--i.e., all components of \( S_{\text{dark}}(\lambda) \) in Eq.~\ref{eq: final model simple}.

To characterize the noise across different acquisition settings, we collected spectra at four integration times: 0.1~s, 0.2~s, 0.5~s, and 1.0~s. For each integration time, 1000 spectra were acquired, resulting in a total of 4000 noise spectra. Of these, 3200 spectra were used for training, 400 for validation, and 400 for testing in our denoising framework. The spectral variance at each wavenumber was computed from this dataset to estimate \( S_{\text{dark}}(\lambda) \), which served as the noise variance term in our simulation model (Eq.~\ref{eq: final model simple}).

\subsection{Generation of Pure Raman Spectra}
\label{sect:pure_raman_generation}

To generate physically realistic pure Raman spectra for simulation, we employed the pseudo-Voigt line shape model, a widely used approximation for modeling Raman peak profiles.~\cite{Beumers2019_PhysBasedRaman, Brubaker2021_RamanCarbonFibers, Korepanov2018_AsymPseudoVoigtRaman} The pseudo-Voigt function is a linear combination of Gaussian and Lorentzian components, offering flexibility to represent Raman peaks with both homogeneous and inhomogeneous broadening characteristics.

Each Raman peak was modeled as:

\begin{equation}
    P(\lambda) = \eta \cdot L(\lambda; \lambda_0, \gamma) + (1 - \eta) \cdot G(\lambda; \lambda_0, \sigma)
\end{equation}

where \( \lambda_0 \) is the center wavenumber (in \( \text{cm}^{-1} \)), \( \gamma \) and \( \sigma \) are the Lorentzian and Gaussian widths, respectively, and \( \eta \in [0, 1] \) is the mixing coefficient controlling the Lorentzian–Gaussian ratio.

For each synthetic Raman spectrum: 1. The number of peaks \( N \) was randomly sampled from a uniform distribution between 0 and 30; 2. Each peak amplitude was sampled uniformly from the range \([0, 1]\); 3. Peak widths were randomly selected from the range 10–200~\(\text{cm}^{-1}\); 4. Peak center positions \( \lambda_0 \) were randomly placed within the valid spectral range.

The complete pure Raman signal was constructed by summing all peak components:

\begin{equation}
    \hat{R}(\lambda) = \sum_{i=1}^{N} P_i(\lambda)
\end{equation}

These simulated spectra reflect the variability of biological Raman signals while being free of fluorescence and sensor-related noise.

\subsection{Fluorescence Baseline Simulation}
\label{sect:fluorescence_simulation}

To simulate the fluorescence background typically observed in Raman spectra of biological tissues, we generated smooth, non-negative baselines using a randomized polynomial approach. These baselines are intended to reflect the low-frequency, continuous background signal that originates from both endogenous fluorophores in the sample and autofluorescence from optical components.

We modeled the fluorescence baseline as a one-dimensional polynomial function of wavenumber \( \lambda \), where the polynomial order and coefficients were sampled randomly to introduce variability across simulated spectra. Specifically, each fluorescence baseline was generated as:

\begin{equation}
    F(\lambda) = \sum_{k=0}^{n} a_k \lambda^k
\end{equation}

where \( n \) is the polynomial order sampled uniformly from the range 3 to 6, and the coefficients \( a_k \) were independently drawn from a uniform distribution in the range \([-1, 1]\). The wavenumber range was set to \( [600, 1790]~\text{cm}^{-1} \), discretized into 693 uniformly spaced points.

To ensure all fluorescence baselines were strictly non-negative across the spectral range, the baseline was either directly accepted if its minimum value was positive, or upward-shifted by subtracting the minimum and adding a small constant (\(10^{-6}\)) if necessary. This process was repeated up to 100 iterations per spectrum to guarantee successful generation of positive-definite curves.

\subsection{Generation of Final Spectra}
\label{sec:generate final spectra}

To simulate realistic Raman spectra with both fluorescence and noise, we first generate clean Raman and fluorescence components as described in previous sections. We then scale them using two target metrics:

\begin{itemize}
    \item \textbf{Raman-to-fluorescence ratio (r2f):} defined as the ratio between the maximum Raman peak amplitude and the maximum fluorescence amplitude:
    \begin{equation}
        \text{r2f} = \frac{\max R(\lambda)}{\max F(\lambda)}
    \end{equation}

    \item \textbf{Signal-to-noise ratio (SNR):} defined as the ratio between the Raman peak amplitude and the noise standard deviation at the same spectral location:
    \begin{equation}
        \text{SNR} = \frac{R(\lambda_p)}{\sqrt{S_{\text{sample}}(\lambda_p) + 2 \cdot S_{\text{dark}}(\lambda_p)}}
    \end{equation}
    where \( \lambda_p \) is the spectral position of the max Raman peak.
\end{itemize}

Given a clean Raman signal \( R(\lambda) \) and a fluorescence baseline \( F(\lambda) \), we determine scale factors \( m \) and \( n \) such that the scaled signals \( \tilde{R}(\lambda) = m \cdot R(\lambda) \) and \( \tilde{F}(\lambda) = n \cdot F(\lambda) \) satisfy the target \( r2f \) and SNR. This $\tilde{R}(\lambda)$ and $\tilde{F}(\lambda)$ is used as the $\hat{R}(\lambda)$ and $\hat{F}(\lambda)$ in Eq.~\ref{eq: final model}. 

For the final spectra, we generated 10000 pure Raman spectra for training, 1000 for validation and 1000 for testing. The number of generated fluorescence baseline is the same. Then, for each training epoch, we randomly select one Raman spectra and one fluorescence baseline, and r2f was uniformly sampled in range [0.1, 0.5], SNR was uniformly sampled in range [0.01, 20], to generate one final spectra.

The derivation of the closed-form solution for \( m \) and \( n \) is provided in Appendix~\ref{appendix:mn_derivation}.

\subsection{Tradition Denoising Method}
\label{sec:traditional denoising}

Before the advent of deep learning, Raman denoising primarily relied on classical signal processing techniques to suppress stochastic noise and remove fluorescence background.

\paragraph{Stochastic Denoising}

For high-frequency noise suppression, two commonly used techniques are the Savitzky–Golay (SG) filter and wavelet denoising.

The SG filter smooths a signal by fitting a local polynomial of degree $d$ within a moving window of length $2m+1$. For a 1D signal $y(x)$, the filtered value at position $x_0$ is given by:
\begin{equation}
    \hat{y}(x_0) = \sum_{i=-m}^{m} c_i y(x_0 + i)
\end{equation}
where $c_i$ are the convolution coefficients determined by least-squares polynomial fitting. This method is effective for preserving peak shapes, but its performance degrades when noise is large relative to the signal or when peak width varies significantly.

Wavelet denoising decomposes a signal using discrete wavelet transform (DWT) into approximation and detail coefficients:
\begin{equation}
    y(t) = \sum_{k} a_{j,k} \phi_{j,k}(t) + \sum_{j < J} \sum_{k} d_{j,k} \psi_{j,k}(t)
\end{equation}
where $\phi$ and $\psi$ are the scaling and wavelet functions at scale $j$, and $a_{j,k}$, $d_{j,k}$ are the corresponding coefficients. Noise is typically concentrated in high-frequency components (detail coefficients), which are attenuated using soft or hard thresholding before signal reconstruction. Although wavelet denoising is more adaptive, it still struggles with mixed-frequency components such as overlapping peaks and high baseline variance.

\paragraph{Baseline Removal}

Fluorescence background is commonly removed using polynomial fitting methods. One widely used technique is the iterative modified polynomial fitting (ModPoly)~\cite{Lieber2003_FluorSub, Zhao2007}, which fits a low-order polynomial to the signal while masking Raman peaks in each iteration. The goal is to fit the baseline curve $b(x)$ such that:
\begin{equation}
    y(x) = s(x) + b(x) + \epsilon(x)
\end{equation}
where $s(x)$ is the true Raman signal and $\epsilon(x)$ is noise.

The algorithm iteratively performs the following steps:
\begin{enumerate}
    \item Fit a polynomial baseline to the masked signal using least squares on the Vandermonde matrix.
    \item Identify regions where the residual $r(x) = y(x) - b(x)$ is positive (i.e., potential peaks).
    \item Replace those regions with baseline estimates and refit.
    \item Repeat until the residual norm change is below a threshold.
\end{enumerate}


In practice, we select the best polynomial order from a predefined range (e.g., 3 to 6) and apply this adaptive baseline correction independently to each Raman spectrum.

While these traditional methods are widely adopted, they often fail when fluorescence intensity is strong or when stochastic noise overlaps with broad baseline features. Our proposed cascaded deep learning framework is designed to address these limitations by explicitly modeling stochastic and baseline components in two separate stages.

\subsection{Deep Learning Model Structure}
\label{sect:deep_learning_model}

\begin{figure}[H]
    \centering
    \includegraphics[width=1\linewidth]{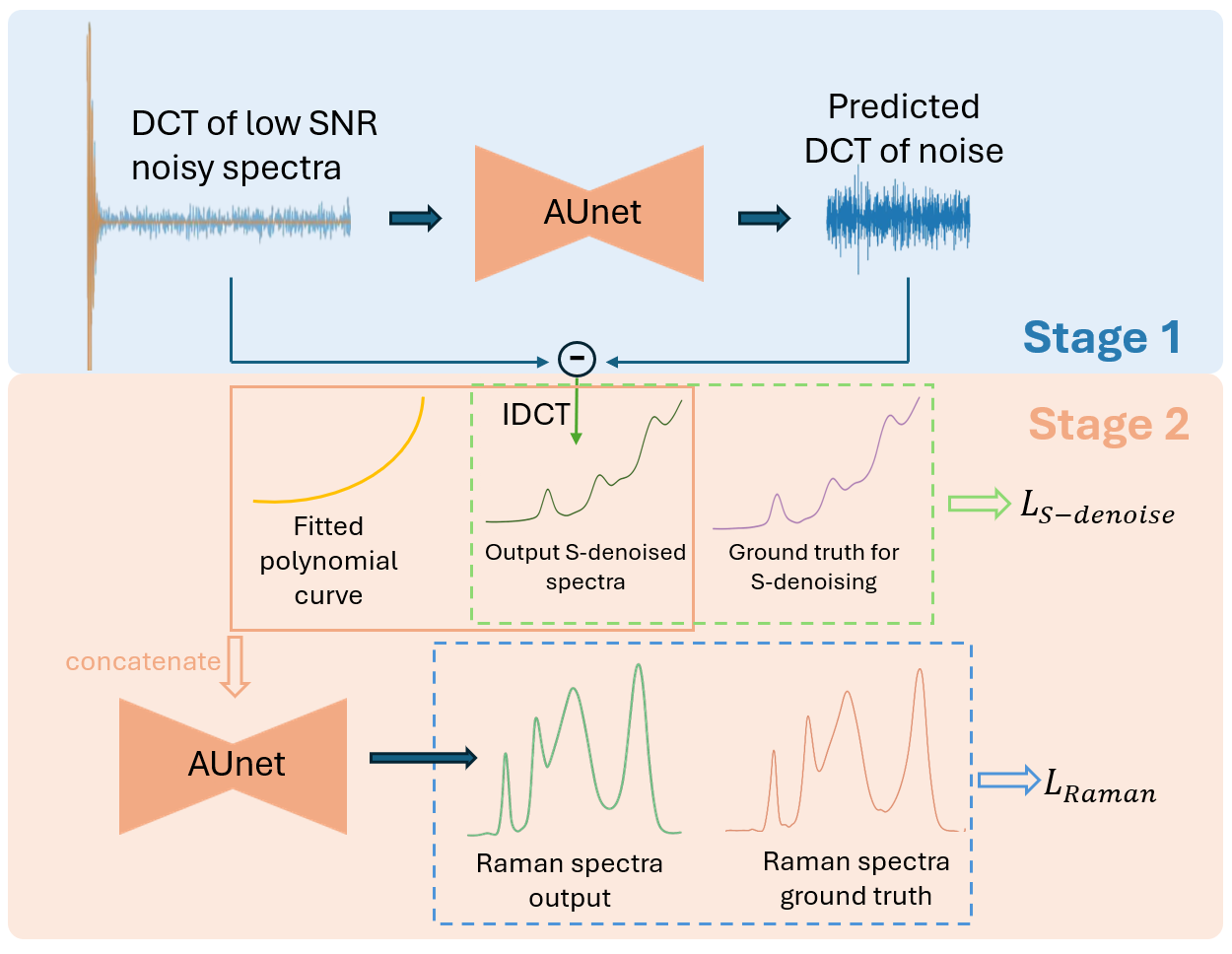}
    \caption{Overview of the dual-stage AUnet-based denoising architecture. The first stage removes stochastic noise via DCT, and the second stage refines the denoised spectrum to match the clean Raman signal.}
    \label{fig:network_structure}
\end{figure}

Our denoising framework is based on a dual-stage cascaded neural network architecture that leverages both spectral-domain and spatial-domain priors. The network is built upon AUnet, an attention-augmented U-Net originally proposed for instrument noise learning in high-contrast hyperspectral microscopy \cite{AUnet_paper}. The overall structure is illustrated in Fig.~\ref{fig:network_structure}.

\paragraph{Stage 1: Frequency-domain stochastic noise estimation}

The first AUnet receives the discrete cosine transform (DCT) of the input low-SNR Raman spectrum and predicts the DCT of the stochastic noise component. This approach is motivated by the observation that Raman peaks tend to be sparse in the frequency domain, whereas stochastic noise (including shot noise, dark noise, and read noise) exhibits broader frequency content.

The predicted DCT noise is subtracted from the DCT of the input signal, and the result is transformed back to the spatial domain using inverse DCT (IDCT). This yields a preliminary denoised spectrum, which we refer to as the “S-denoising” output, to distinguish it from the overall denoised result. A fitted polynomial baseline using Section \ref{sec:traditional denoising} baseline removal is then computed and concatenated with the S-denoised spectrum to form the input to the second stage.

\paragraph{Stage 2: Baseline removal and Raman spectra recovery}

The second AUnet receives the concatenated S-denoised signal and fitted polynomial curve as input. It focuses on removing the fluorescence baseline and refining Raman peak structures. This stage is designed to correct low-frequency distortions and enhance fine spectral details that remain after stochastic denoising.

\paragraph{Training Objectives}

The network is supervised using two loss functions:

\begin{itemize}
    \item \( \mathcal{L}_{\text{S-denoise}} \): mean squared error (MSE) between the S-denoised output (after IDCT) and the clean baseline-included ground truth spectrum.
    \item \( \mathcal{L}_{\text{Raman}} \): MSE between the final network output and the baseline-free pure Raman signal.
\end{itemize}

The total loss is computed as a weighted sum:

\begin{equation}
    \mathcal{L}_{\text{total}} = \alpha \cdot \mathcal{L}_{\text{S-denoise}} + (1 - \alpha) \cdot \mathcal{L}_{\text{Raman}}
\end{equation}

This cascaded design allows the model to first isolate stochastic noise in the frequency domain and then remove the fluorescence baseline in the spatial domain, resulting in a clean and interpretable Raman spectrum.

\paragraph{Training Procedure}

We adopt a three-phase training strategy to stabilize learning across both stages:
\begin{enumerate}
    \item First, we fix \( \alpha = 1 \) to train only the S-denoising stage and freeze the weights for the second AUnet for 50 epochs using a batch size of 32 and a learning rate of \( 2 \times 10^{-5} \).
    \item Next, we set \( \alpha = 0 \) to train the second stage exclusively and freeze the weights for the first AUnet for 100 epochs with the same hyperparameters.
    \item Finally, we jointly fine-tune the entire network by setting \( \alpha = 0.5 \) and training for an additional 50 epochs.
\end{enumerate}

\subsection{Human Skin Spectra Generation}
\label{sec: skin_spectra_generation}

Simulated human skin Raman spectra were generated based on the biophysical modeling framework described in Feng \textit{et. al}'s~\cite{Xu_biophysical} work. In that work, Raman spectra were collected for seven main biophysical components of human skin: water, ceramide, keratin, nucleus, triolein, elastin, and collagen, which were treated as biophysical origins.

To ensure comparability and preserve relative peak structures, each of these component spectra was first normalized by the area under the curve (AUC). This normalization preserves the relative shape and positions of spectral features while allowing for flexible amplitude scaling. 

In our simulation, we treated these seven normalized spectra as basis functions and generated synthetic skin spectra by performing a weighted linear combination:

\begin{equation}
    R_{\text{skin}}(\lambda) = \sum_{i=1}^{7} w_i \cdot C_i(\lambda)
\end{equation}

where \( C_i(\lambda) \) is the normalized spectrum of the \( i \)-th biophysical component, and \( w_i \sim \mathcal{U}(0, 1) \) is a randomly sampled weight for each spectrum. This formulation reflects biological variability across different skin locations and individuals by simulating variation in biochemical composition.

A total of 1000 skin spectra were generated for testing. These simulated spectra serve as the clean Raman signal \( \hat{R}(\lambda) \) in our denoising framework, allowing us to benchmark model performance under controlled conditions with known ground truth.

Because the simulated spectra represent only the Raman scattering contribution of skin (i.e., they are free from fluorescence and noise), they play a central role in evaluating both the stochastic denoising and fluorescence removal capabilities of our proposed method.

\subsection{Evaluations}
\subsubsection{Evaluation of SNR Improvement After S-denoising}
\label{subsec:snr_improvement}

To quantitatively evaluate the effect of the first-stage denoising network (S-denoising), we designed a comprehensive simulation protocol to measure the SNR improvement across a range of noise conditions.

We began by generating 500 random pairs of $(\text{r2f}, \text{SNR})$ values to span a wide spectrum of Raman-to-fluorescence ratios and noise levels. For each of these 500 pairs (indexed by $i$), we randomly selected 5 clean Raman spectra and 5 clean fluorescence backgrounds from the 1000 available test spectra, and combined them to form 5 clean composite spectra, denoted as $S_{i,j}$, where $j = 1, 2, \dots, 5$.

To simulate measurement variability, each clean spectrum $S_{i,j}$ was corrupted by stochastic noise 10 times, resulting in 10 distinct low-SNR noisy spectra for each clean signal. These noisy versions are denoted as $S_{i,j,k}$, where $k = 1, 2, \dots, 10$. The stochastic noise was generated to match the target SNR value specified by the pair $(\text{r2f}_i, \text{SNR}_i)$.

Each of the resulting $500 \times 5 \times 10 = 25{,}000$ noisy spectra $S_{i,j,k}$ was then processed through the first-stage denoising model, yielding outputs $O_{i,j,k}$.

To evaluate the noise level after S-denoising, we used the following procedure:
\begin{enumerate}
    \item Identify the maximum Raman peak position $x_p$ from the clean signal $S_{i,j}$.
    \item At this position $x_p$, compute the standard deviation of the denoised outputs across the $k = 1 \dots 10$ realizations:
    \[
    \sigma_{i,j} = \text{std}_k \left( O_{i,j,k}(x_p) \right)
    \]
    \item The corresponding signal level is the amplitude at $x_p$ from the clean signal: $A_{i,j} = S_{i,j}(x_p)$.
    \item Compute the SNR after denoising as:
    \[
    \text{SNR}_{i,j}^{\text{new}} = \frac{A_{i,j}}{\sigma_{i,j}}
    \]
    \item Compute the SNR improvement using the logarithmic metric:
    \[
    \text{SNRi}_{i,j} = 10 \cdot \log_{10} \left(1 + \frac{\text{SNR}_{i,j}^{\text{new}} - \text{SNR}_i^{\text{old}}}{\text{SNR}_i^{\text{old}}} \right)
    \]
\end{enumerate}

Finally, the overall SNR improvement for each $(\text{r2f}_i, \text{SNR}_i)$ pair is calculated by averaging the improvements over all 5 clean signals:
\[
\text{SNRi}_i = \frac{1}{5} \sum_{j=1}^{5} \text{SNRi}_{i,j}
\]

We compare the SNR improvement of our deep learning-based method with the Savitzky-Golay (SG) filter and the Wavelet filter mentioned in Section.~\ref{sec:traditional denoising}.

\subsubsection{Raman Peak Analysis}
\label{subsec:peak_analysis}

In addition to evaluating overall signal-to-noise ratio (SNR) improvement, we further assessed the denoising performance of our model by analyzing its ability to accurately recover Raman peaks. For this experiment, we generated 1000 synthetic test spectra by randomly selecting from the 1000 available pure Raman signals and 1000 fluorescence backgrounds. For each generated spectrum, random $(\text{r2f}, \text{SNR})$ values were sampled, following the same protocol as described in the training and SNR evaluation sections.

The denoised outputs were obtained from the second-stage AUnet in our cascaded architecture. We then applied the \texttt{scipy.signal.find\_peaks} function to detect peaks in the denoised spectra. To evaluate robustness across different peak intensities, we varied the \texttt{prominence} parameter to control the sensitivity of peak detection. A higher prominence value filters out weaker peaks, retaining only the dominant ones, whereas lower values include more subtle peaks.

To quantify the accuracy of peak recovery, we compared the detected peaks in the denoised spectra to those in the corresponding ground truth clean Raman signals. A detected peak was considered a match if its wavenumber position deviated from a true peak by no more than $\pm$6~cm$^{-1}$.

The following metrics were computed for each test spectrum:

\begin{itemize}
    \item \textbf{Missing peak ratio:}
    \begin{equation}
        \text{Missing Ratio} = \frac{N_{\text{miss}}}{N_{\text{true}}}
    \end{equation}
    where $N_{\text{miss}}$ is the number of true peaks not matched by any predicted peak, and $N_{\text{true}}$ is the total number of true peaks.

    \item \textbf{Artifact peak ratio:}
    \begin{equation}
        \text{Artifact Ratio} = \frac{N_{\text{artifact}}}{N_{\text{true}}}
    \end{equation}
    where $N_{\text{artifact}}$ is the number of predicted peaks not matched to any true peak.

    \item \textbf{Peak value bias:}
    \begin{equation}
        \text{Peak Value Bias} = \frac{1}{N_{\text{match}}} \sum_{i=1}^{N_{\text{match}}} \left| A_i^{\text{pred}} - A_i^{\text{true}} \right|
    \end{equation}
    where $A_i^{\text{pred}}$ and $A_i^{\text{true}}$ are the amplitudes of the $i$-th matched peak in the denoised and ground truth spectra, respectively.

    \item \textbf{Peak shift:}
    \begin{equation}
        \text{Peak Shift} = \frac{1}{N_{\text{match}}} \sum_{i=1}^{N_{\text{match}}} \left| \lambda_i^{\text{pred}} - \lambda_i^{\text{true}} \right|
    \end{equation}
    where $\lambda_i^{\text{pred}}$ and $\lambda_i^{\text{true}}$ are the wavenumber positions of the $i$-th matched peak in the denoised and ground truth spectra, respectively.
\end{itemize}

We compared the peak performance with the spectra after fluorescence removal using polynomial fit in Section.~\ref{sec:traditional denoising}. This evaluation provides a detailed understanding of the model’s ability to recover the structural and quantitative features of Raman spectra, beyond just global denoising performance.

\subsubsection{Human Skin Spectra Analysis}
\label{sec: huamn skin gen}
In addition to evaluating the SNR improvement and peaks, it is crucial to ensure that the Raman spectra patterns remain accurate after denoising. As mentioned in Section.~\ref{sec: skin_spectra_generation}, the final skin spectra were generated by assigning random concentrations ranging from $0$ to $1$ to seven biophysical components. To evaluate the accuracy of the spectra patterns, NNLS was used to predict the concentrations from both the pure skin spectra and the denoised spectra. It is important to note that a direct comparison with the originally assigned concentrations is not appropriate because we have scaled the pure spectra to match the target SNR value when generating the low-SNR spectra, and NNLS also inherently introduces bias. Therefore, NNLS was applied to both the pure spectra and the denoised spectra, allowing for a direct comparison.
        
We plotted the concentrations from the denoised spectra against the concentrations from the pure spectra and fitted a linear curve. If the denoised spectra preserve the spectral information accurately, the fitted line should closely match the diagonal line: 

\begin{equation}
    C_{i}^{pred} = C_i
\end{equation}

where $C_i$ is the absolute concentration of component $i$ and $C_{i}^{pred}$ is the predicted concentration of component $i$.

Additionally, we reported the MSE value between the two concentrations to quantitatively evaluate the concentration accuracy and compared the deep learning model output with the traditional method.

\section{Results}

\subsection{Spectra Simulation}
To train and evaluate our denoising model, we constructed a synthetic dataset of Raman spectra corrupted by varying levels of stochastic noise and fluorescence background. Each simulated spectrum is a combination of a randomly generated pure Raman signal and a fluorescence background, with noise components added according to the sensor characteristics.

The pure Raman signals were generated using the pseudo-Voigt method, simulating realistic peak shapes by sampling the number of peaks (0–30), peak positions, amplitudes (uniformly from 0 to 1), and full width at half maximum (FWHM, from 10 to 200$cm^{-1}$). The fluorescence background was modeled using a random polynomial curve of order 3–6 and scaled according to a target R2F. Stochastic noise was then added to the combined signal based on a specified SNR level, using a mixture of Poisson and Gaussian distributions to mimic real-world sensor noise, including shot noise, dark noise, and read noise.

\begin{figure}[htbp]
    \centering
    \includegraphics[width=\linewidth]{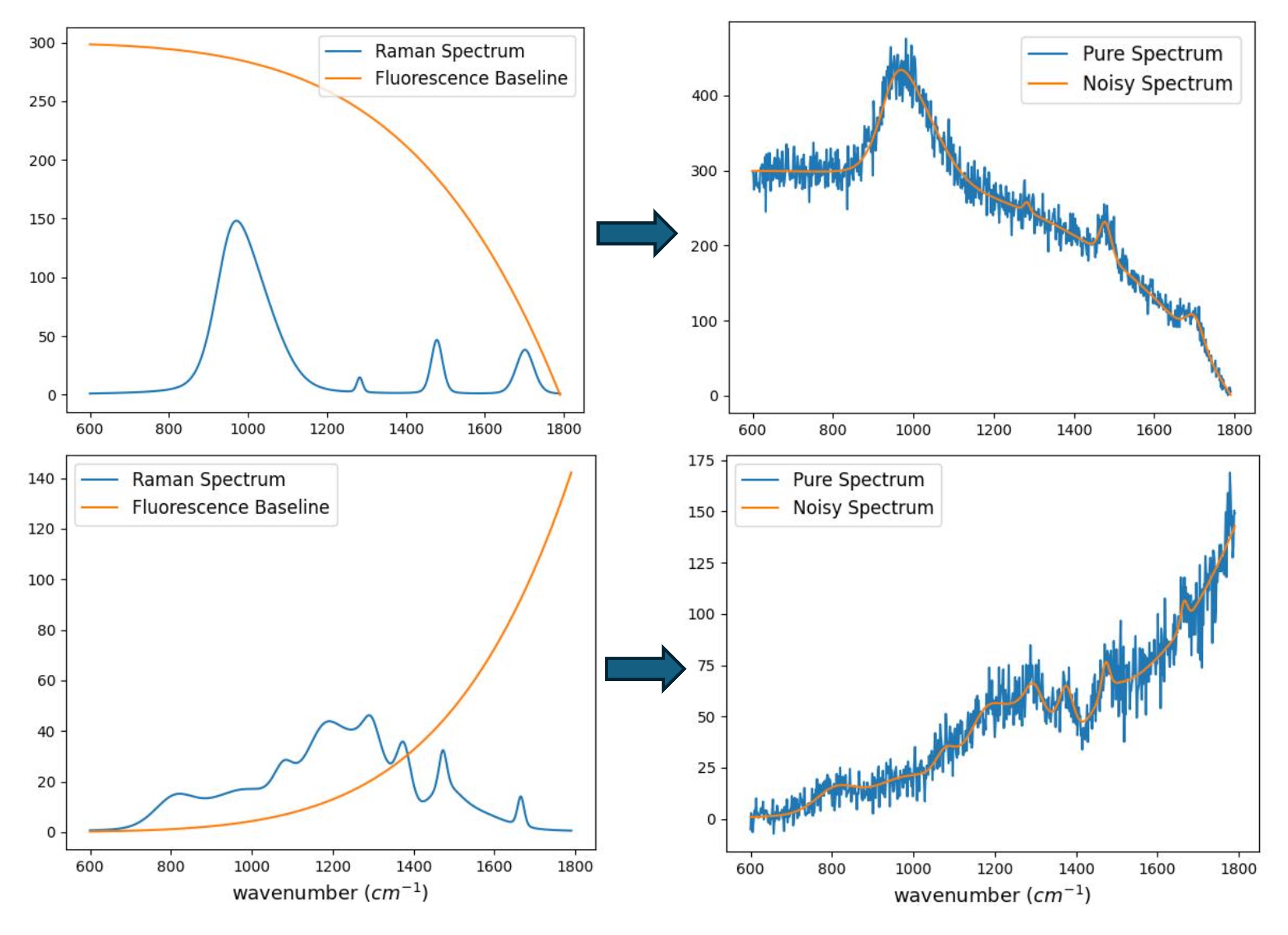}
    \caption{
        Simulated spectra examples. Left columns are two examples of pure Raman spectrum and fluorescence spectrum examples. Right columns are the combination the Raman and the fluorescence with the noise added.
    }
    \label{fig:spectrum_example}
\end{figure}

As illustrated in Fig.~\ref{fig:spectrum_example}, the resulting simulated spectrum realistically reflects the challenges encountered in practice: Raman peaks are partially buried under broad fluorescence baselines and noise artifacts. By varying the r2f and SNR levels systematically, this dataset enables our model to learn robust denoising behavior across a wide range of spectral qualities. More examples of simulated spectra can are shown in Appendix~\ref{append: raman and fluorescence}.

\subsection{SNR Improvement}
\label{sect:results_snri}

To validate the effectiveness of the first-stage stochastic denoising, we evaluate the improvement in SNR achieved after denoising. A high SNR is essential for reliable Raman peak detection and accurate spectral analysis. We compare our proposed deep learning method against two commonly used traditional denoising approaches: the SG filter and the wavelet filter.

\begin{figure}[htbp]
    \centering
    \includegraphics[width=\linewidth]{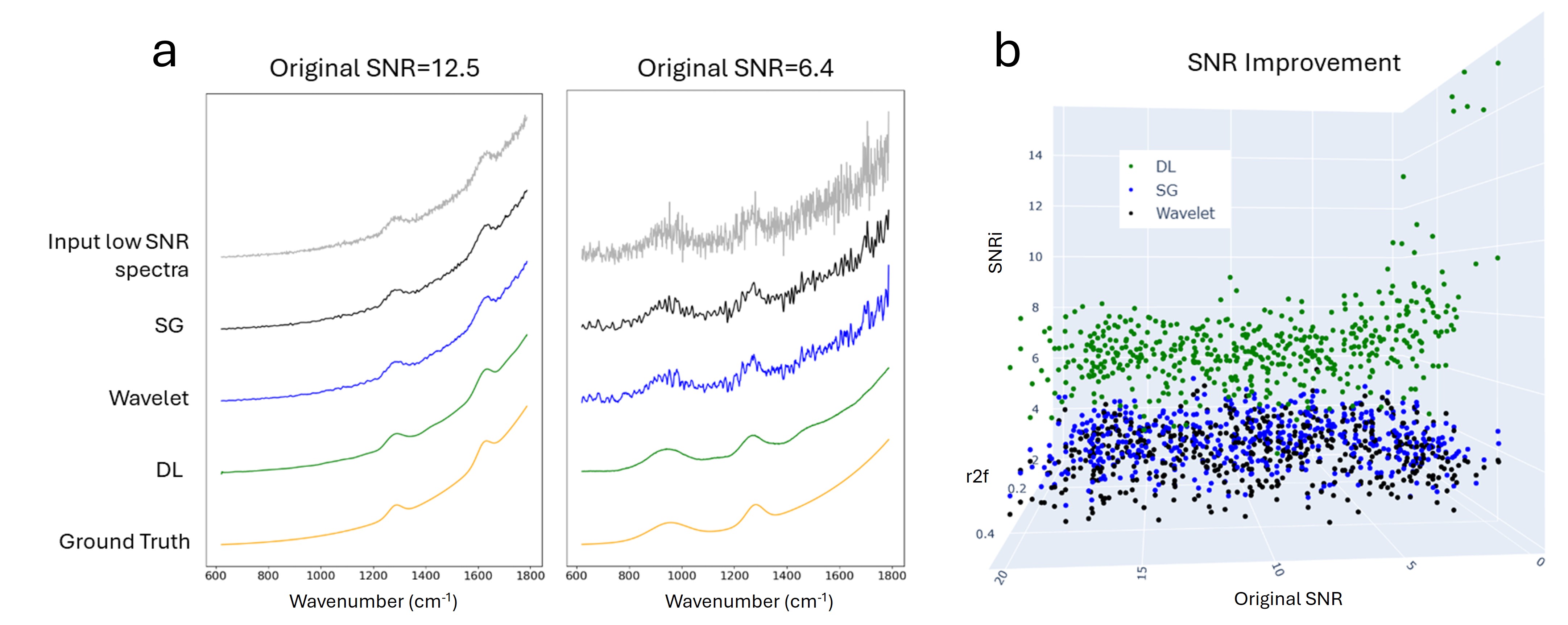}
    \caption{
        \textbf{SNR improvement comparison.} 
        (\textbf{a}) Example Raman spectra before and after denoising using Savitzky-Golay (SG) filter, wavelet filter, and the proposed deep learning (DL) model. Two representative input spectra with initial SNRs of 12.5 and 6.4 are shown. DL achieves a visually smoother result that more closely matches the ground truth.
        (\textbf{b}) Quantitative comparison of SNR improvement across 500 $(\text{r2f}, \text{SNR})$ pairs. Each point represents the average improvement over 5 signals per pair. DL consistently outperforms SG and wavelet filters across different noise and fluorescence conditions.
    }
    \label{fig:snri_comparison}
\end{figure}

The results are summarized in Fig.~\ref{fig:snri_comparison}.

Fig.~\ref{fig:snri_comparison} illustrates both qualitative and quantitative comparisons of SNR improvement across different denoising methods. In panel (a), we show two representative spectra with original SNRs of 12.5 and 6.4, respectively. The input noisy spectra are shown alongside the results from SG, wavelet filtering, and our DL method, with the clean ground truth provided for reference. As observed, SG and wavelet methods reduce high-frequency noise to some extent, but often leave behind residual artifacts. In contrast, our DL model more effectively preserves spectral peak structure while substantially suppressing stochastic noise.

Panel (b) presents a quantitative comparison of SNR improvement for 500 randomly selected $(\text{r2f}, \text{SNR})$ pairs. For each pair, 5 clean Raman signals were simulated and corrupted with 10 realizations of stochastic noise, followed by denoising using each method. The resulting SNR improvements were averaged per pair and plotted in a 3D scatter plot with axes representing original SNR, r2f ratio, and SNR improvement. As shown, the DL method consistently outperforms SG and wavelet filters across the entire spectrum of input conditions. The SNR gains achieved by DL are not only higher on average but also more robust across both low and high fluorescence backgrounds, validating its generalizability and effectiveness for Raman denoising.

\subsection{Raman Peak Evaluation}
\label{sect:results_peak_eval}

While many Raman analysis tasks focus on detecting a few specific biomarker peaks, evaluating denoising performance solely on those peaks may not reflect generalizability across different biological conditions. In contrast, for denoising model validation, it is important to assess how well the full spectrum is preserved—including both strong and weak peaks—under varying levels of noise and fluorescence background. To this end, we systematically evaluated peak recovery accuracy across the entire spectrum for the second AUnet outputs using a set of objective metrics, as described in Section~\ref{subsec:peak_analysis}.

\begin{figure}[H]
    \centering
    \includegraphics[width=0.8\linewidth]{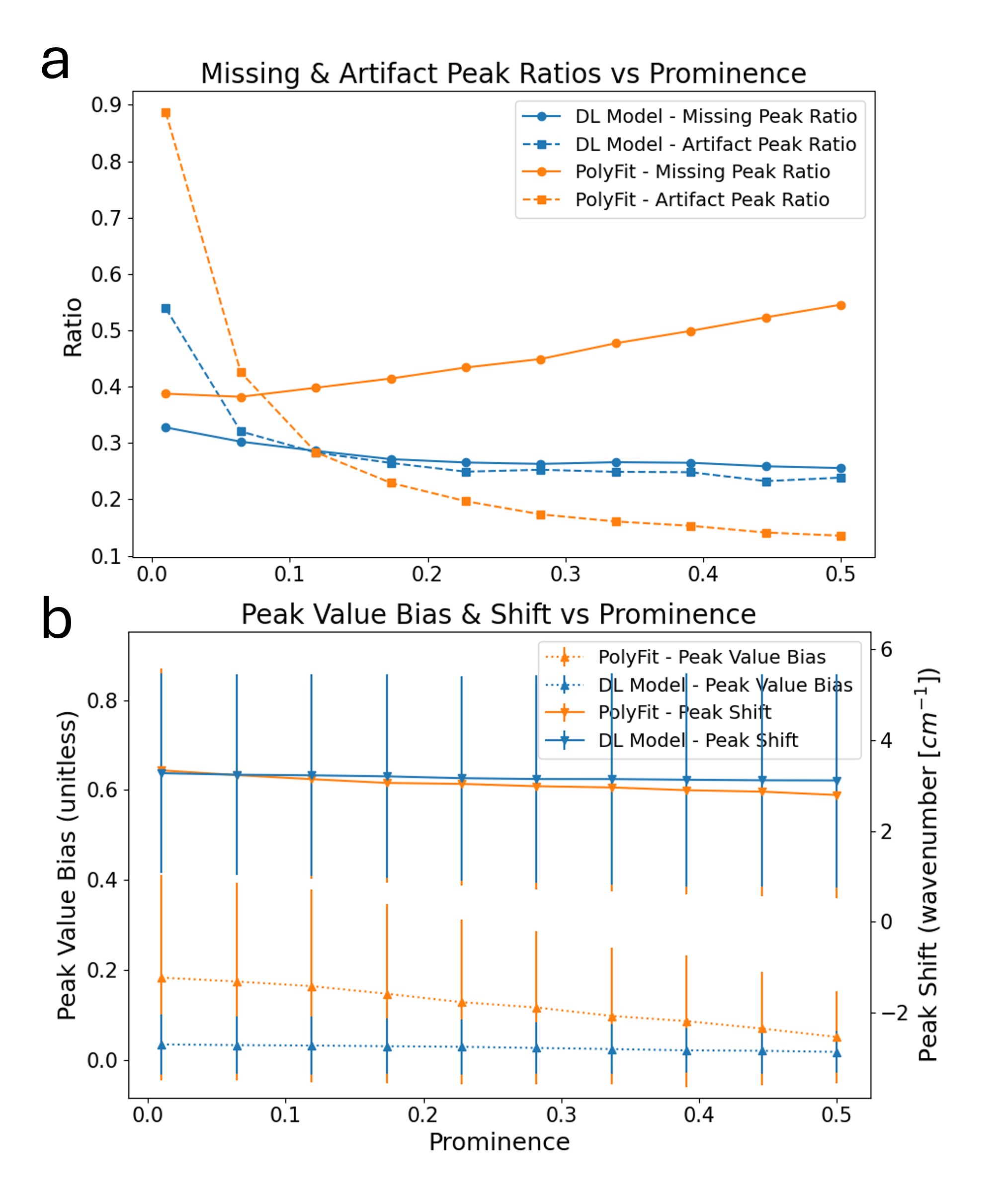}
    \caption{
        \textbf{Raman peak evaluation across prominence thresholds.}
        (\textbf{a}) Missing peak ratio and artifact peak ratio for DL and PolyFit models. DL consistently produces fewer false negatives across all prominence levels. But for the false positives, becaus
        (\textbf{b}) Peak value bias and peak shift of matched peaks. DL achieves lower intensity bias. Error bars represent standard deviation.
    }
    \label{fig:peak_eval}
\end{figure}

Fig.e~\ref{fig:peak_eval} summarizes the peak analysis results comparing our deep learning model with a traditional baseline removal method based on polynomial fitting (PolyFit). Performance was evaluated across a range of peak prominence thresholds to test the model’s ability to recover both subtle and dominant spectral features.

In panel (a), we observe that the DL model consistently achieves lower missing and artifact peak ratios across all prominence levels. In contrast, the PolyFit method exhibits a sharp increase in missing peak ratio as the prominence threshold increases. This suggests that while PolyFit can preserve prominent peaks to some degree, it quickly fails to recover weaker peaks as the detection threshold rises. Interestingly, the artifact peak ratio for PolyFit decreases at high prominence, but this reduction may be misleading. A likely explanation is that as PolyFit misses more true peaks, the denoised spectra become increasingly flat, leaving fewer detectable peaks overall—thus reducing the denominator in the artifact ratio calculation and artificially lowering its value.

Panel (b) illustrates that the DL model achieves significantly lower peak value bias compared to PolyFit. The DL model maintains consistent performance across all prominence levels, while PolyFit’s errors in peak intensity gradually worsen, especially in low-prominence regimes. The two models have similar peak shift performance.

These results demonstrate that our model not only reduces noise effectively but also preserves the detailed spectral structure needed for robust Raman interpretation across a wide range of signal conditions.

\subsection{Human skin spectra analysis}
To further validate the practical applicability of our denoising framework, we tested its performance on simulated human skin spectra. Human skin serves as a compelling use case due to its complex biochemical composition and strong endogenous fluorescence, which pose significant challenges for Raman analysis. In biomedical applications such as skin cancer diagnosis or tissue margin assessment, accurately resolving Raman features from skin is critical. Therefore, using human skin as an example allows us to demonstrate how well our model generalizes to realistic, heterogeneous biological signals under severe fluorescence interference and stochastic noise.

To generate realistic human skin spectra, we employed a real-data-based simulation approach grounded in the biophysical skin model developed by Feng et al.~\cite{Xu_biophysical}. In that study, Raman spectra were experimentally collected from seven key skin components: water, ceramide, keratin, nucleus, triolein, elastin, and collagen. These components were identified as the primary biochemical contributors to skin’s Raman signature. Each component spectrum was normalized by its area under the curve to ensure consistent scaling. We then randomly assigned non-negative weights to each component and computed a weighted linear combination to simulate composite skin spectra. This approach captures the natural variability in skin tissue composition and enables the generation of a large, diverse dataset for testing denoising performance in a biologically relevant context.

We then used 1000 simulated clean skin spectra and assigned each spectrum a randomly selected (r2f,SNR) pair. The final noisy skin spectra were generated by combining the clean signal with a simulated fluorescence baseline and adding stochastic noise using the same statistical model described in Section~\ref{sect:noise model}. This procedure yielded a diverse and biologically realistic test set, representative of the range of noise and fluorescence conditions encountered in real skin Raman measurements. the simulated skin spectra examples are shown in Appendix~\ref{append: skin}.

\begin{figure}[H]
\centering
\includegraphics[width=\linewidth]{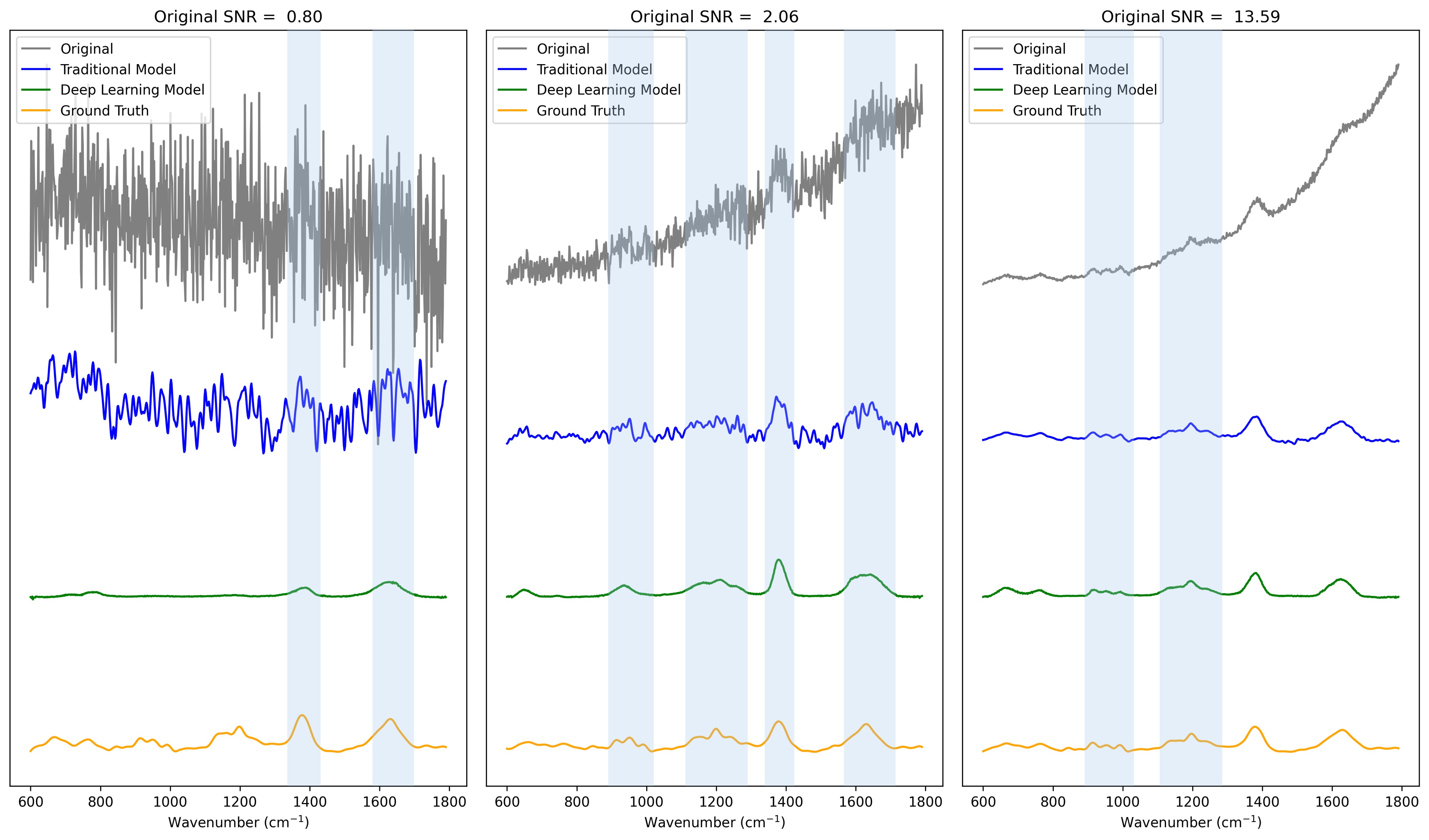}
\caption{
\textbf{Denoising results on simulated human skin spectra at different SNR levels.}
Each panel shows the original noisy input (gray), traditional polynomial baseline result (blue), our DL model output (green), and ground truth (orange). Highlighted regions indicate Raman bands of interest.}
\label{fig:skin_example}
\end{figure}

To qualitatively assess the performance of our model on biologically realistic data, Figure~\ref{fig:skin_example} presents three representative examples of denoised human skin spectra at low, medium, and high SNR levels. Each subplot shows the original noisy spectrum, the result of traditional polynomial baseline correction, the output from our DL model, and the clean ground truth.

At low SNR (left panel, SNR=0.80), our DL model successfully recovers major peaks corresponding to key biochemical markers—particularly the lipid band around 1400cm$^{-1}$ and protein band near 1600cm$^{-1}$—whereas the traditional method fails to suppress noise and loses most spectral structure. In the medium SNR case (middle panel, SNR=2.06), both methods reduce noise, but the DL model produces a cleaner signal and preserves more minor peaks and finer spectral features. At high SNR (right panel, SNR=13.59), both approaches perform well overall. However, in some spectral regions (e.g., the highlighted blue bands), the traditional method exhibits slightly more accurate reconstruction, possibly due to its tendency to follow smoother baselines at low noise levels.

These examples demonstrate that our model is particularly effective in low- and mid-SNR conditions where traditional methods often fail, while still performing comparably well in high-SNR regimes.

To further assess the practical utility of our denoised spectra, we evaluated how well they preserve underlying biochemical information by decomposing the spectra back into constituent skin components. Specifically, we applied non-negative least squares (NNLS) to estimate the concentration of each biophysical component using the denoised spectra and the known reference signatures of collagen, elastin, triolein, nucleus, keratin, ceramide, and water.

\begin{figure}[H]
\centering
\includegraphics[width=\linewidth]{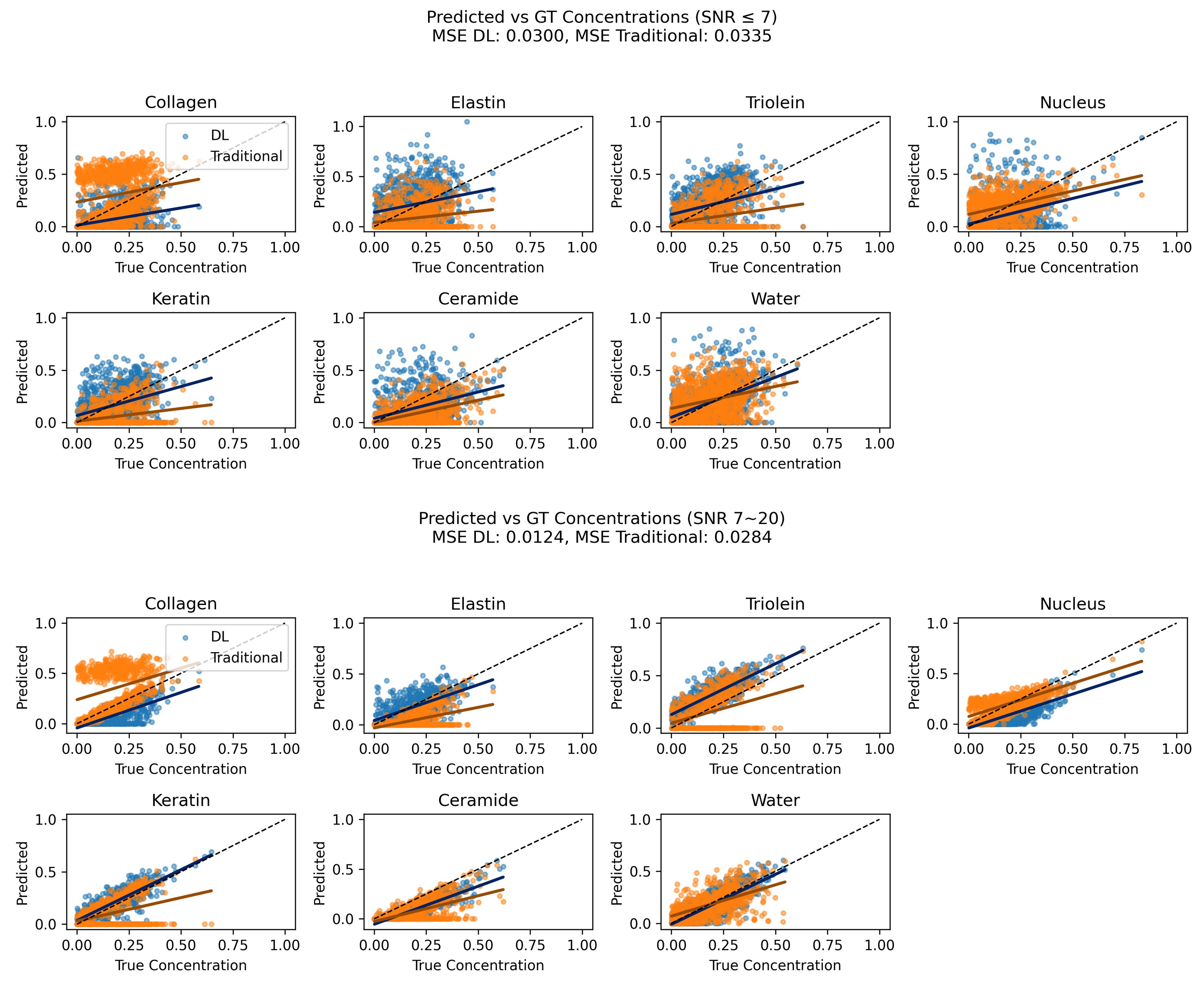}
\caption{
\textbf{Predicted vs. ground truth concentrations of skin components using NNLS decomposition.}
Denoised spectra from the DL model (blue) and the traditional method (orange) are used to estimate component concentrations for low SNR (top) and high SNR (bottom) conditions. Solid lines are linear fits. The DL model achieves significantly lower mean squared error (MSE) and better concentration recovery, particularly in the low-SNR regime.
}
\label{fig:concentration_prediction}
\end{figure}

Figure~\ref{fig:concentration_prediction} shows the predicted concentrations plotted against the ground truth for both DL model and the traditional polynomial baseline method. Results are shown separately for low-SNR spectra (SNR$\leq$7, top panel) and high-SNR spectra (7$<$SNR$\leq$20, bottom panel). Each subplot corresponds to one component, and linear fits are overlaid to show the trend.

In the low-SNR regime, our DL model provides substantially better recovery of component concentrations, with closer alignment to the 1:1 identity line and lower prediction variance. The fitted lines for DL (blue) consistently show stronger slopes and better correlation with the ground truth compared to the traditional approach. Quantitatively, the DL model achieves a lower mean squared error (MSE=0.0300) versus the traditional model (MSE=0.0335). The difference becomes more pronounced in the high-SNR range, where the DL method further reduces its MSE to 0.0124, while the traditional method still lags at 0.0284.

These results demonstrate that our denoising model not only improves spectral quality but also enhances the downstream quantitative analysis of biochemical composition, which is critical for clinical interpretation.

\section{Discussion}
While our results demonstrate that the proposed deep learning framework effectively denoises Raman spectra and preserves key biochemical information, there are several limitations that warrant discussion and opportunities for future improvement.

First, our study was conducted entirely on simulated data. Although the simulation pipeline was carefully designed using real biophysical components and physically grounded noise models, real-world data may present additional challenges, such as unmodeled background signals, system-specific artifacts, or subtle variations in sample composition. The model's generalization to real experimental spectra, especially from clinical instruments or in vivo environments, remains to be validated. Future work will involve domain adaptation or fine-tuning with a small amount of real-world data to bridge this gap.

Second, our noise model was derived based on four discrete integration times. During data generation, each spectrum was randomly assigned one of these integration times, and the noise variance was scaled accordingly. While our current formulation includes integration time implicitly in the variance term and we assume the model is robust to this randomness, this assumption needs further validation. Future extensions could involve explicitly incorporating integration time as an additional input to the model or using a continuous formulation to better generalize across all exposure durations.

Third, we assumed that fluorescence and Raman contributions are additive and uncorrelated. While this is a common and reasonable approximation in the literature, in real systems there may be spectral regions where scattering and fluorescence interact in complex ways due to resonance effects or optical interference. Our current model does not explicitly account for such nonlinear interactions.

Additionally, our cascaded model was trained in a supervised fashion using clean spectra as ground truth. In many practical settings, acquiring clean spectra is expensive or infeasible. To address this limitation, future work may explore self-supervised or weakly supervised training strategies using only partially labeled or unlabeled data. Physics-informed networks or hybrid models that incorporate instrument priors and noise statistics explicitly into the training loss could further enhance robustness.

Lastly, while we demonstrated strong performance on skin-relevant spectra, further evaluation is needed across other biological tissues, chemicals, or use cases such as biofluids, pharmaceutical formulations, or food safety inspection. Generalizing to these domains will require extending our simulation framework to capture domain-specific spectral profiles and background behaviors.

In summary, although our method shows promising results in controlled simulated conditions, expanding its applicability to real-world Raman spectroscopy will require additional work in model generalization, calibration, and validation across broader datasets and acquisition parameters.

\section{Conclusion}
\label{sect:conclusion}

In this study, we proposed a cascaded deep learning framework for denoising Raman spectra, specifically designed to address the challenges posed by low signal-to-noise ratios and strong fluorescence background—conditions commonly encountered in biological tissue analysis. By developing a statistically grounded noise model that incorporates photon shot noise, sensor-specific stochastic components, and fluorescence interference, we simulated realistic Raman spectra with controllable SNR and fluorescence ratios. Our two-stage AUnet architecture effectively decouples stochastic noise suppression and baseline fluorescence removal, enabling robust signal reconstruction across a wide range of conditions.

Through systematic evaluations—including SNR improvement, peak recovery accuracy, and biophysical concentration prediction—we demonstrated that our model consistently outperforms traditional methods such as Savitzky-Golay filtering and polynomial baseline subtraction. Importantly, we validated the method on simulated human skin spectra, showing that it preserves key biochemical features relevant to downstream interpretation tasks.

Overall, this work provides a practical and extensible solution for Raman denoising in complex biological contexts. Future directions include validating the model on real experimental data, generalizing to arbitrary integration times, and extending its use to other spectroscopic and biomedical applications.

\appendix    

\section{Derivation of Signal Scaling Parameters}
\phantomsection
\label{appendix:mn_derivation}

To satisfy both the desired Raman-to-fluorescence ratio \( r2f \) and the target signal-to-noise ratio \( \text{SNR} = s \), we define:

\begin{itemize}
    \item \( x_p \): the amplitude of the maximum Raman peak
    \item \( f_{\max} \): the maximum amplitude of the fluorescence baseline
    \item \( f_p \): the fluorescence amplitude at the max Raman peak location
    \item \( y \): the noise variance for dark frame, approximated as \(2 \cdot S_{\text{dark}} \) at the max Raman peak location
\end{itemize}

We want the scaled signals \( \tilde{R} = m \cdot R \) and \( \tilde{F} = n \cdot F \) to satisfy:

\begin{equation}
    \frac{m \cdot x_p}{n \cdot f_{\max}} = r2f
    \quad \text{and} \quad
    \frac{m \cdot x_p}{\sqrt{m \cdot x_p + n \cdot f_p + y}} = s
\end{equation}

Solving the first equation for \( m \):

\begin{equation}
    m = \frac{r2f \cdot n \cdot f_{\max}}{x_p}
\end{equation}

Substitute into the second equation:

\begin{equation}
    \frac{r2f \cdot n \cdot f_{\max}}{\sqrt{r2f \cdot n \cdot f_{\max} + n \cdot f_p + y}} = s
\end{equation}

Square both sides:

\begin{equation}
    \frac{(r2f \cdot n \cdot f_{\max})^2}{r2f \cdot n \cdot f_{\max} + n \cdot f_p + y} = s^2
\end{equation}

Rewriting as a quadratic equation:

\begin{equation}
    A n^2 + B n + C = 0
\end{equation}

where:

\[
A = (r2f)^2 \cdot f_{\max}^2, \quad
B = -s^2 \cdot (r2f \cdot f_{\max} + f_p), \quad
C = -s^2 \cdot y
\]

Solving for \( n \):

\begin{equation}
    n = \frac{-B + \sqrt{B^2 - 4AC}}{2A}
\end{equation}

and

\begin{equation}
    m = \frac{r2f \cdot n \cdot f_{\max}}{x_p}
\end{equation}

This ensures that the simulated spectra match the desired noise and signal balance conditions.


\section{Spectra Examples}
\subsection{Simulated Raman and Fluorescence Spectra}
\phantomsection
\label{append: raman and fluorescence}

\begin{figure}[H]
\centering
\includegraphics[width=\linewidth]{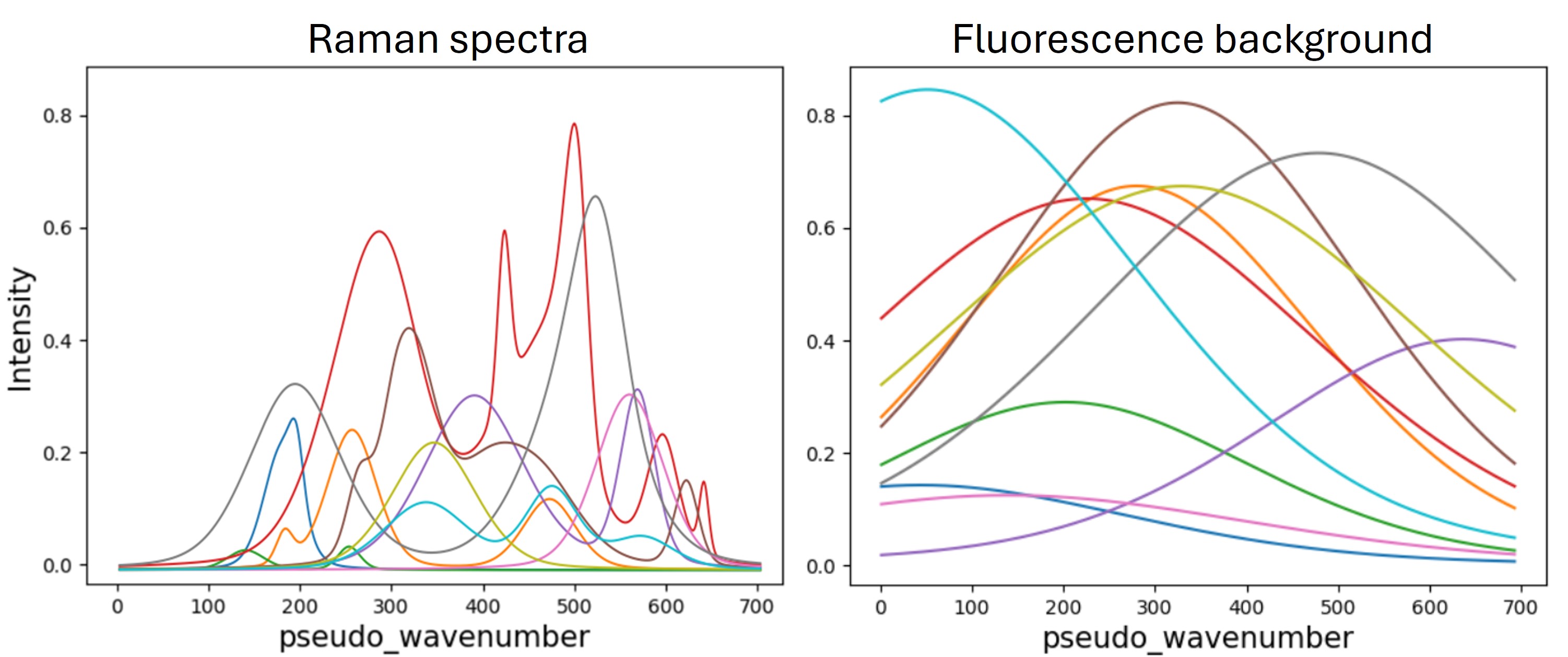}
\caption{
\textbf{Examples of simulated Raman spectra and fluorescence backgrounds.} Each plot is normalized to $[0,1]$ for visualization, and the x-axis represents pseudo wavenumber (data index).
}
\label{fig:spectrum_example_appendix}
\end{figure}

Figure~\ref{fig:spectrum_example_appendix} presents ten examples of simulated Raman spectra (left) and fluorescence background signals (right). Since the simulated data span different amplitude ranges, each spectrum is normalized to the range $[0, 1]$ for visualization purposes. Note that wavenumber values were not explicitly assigned during the simulation process; therefore, the x-axis is labeled as "pseudo wavenumber" to indicate data point indices rather than physical units.

\subsection{Simulated Human Skin Spectra and Biophysical Basis}
\phantomsection
\label{append: skin}

\begin{figure}[H]
\centering
\includegraphics[width=\linewidth]{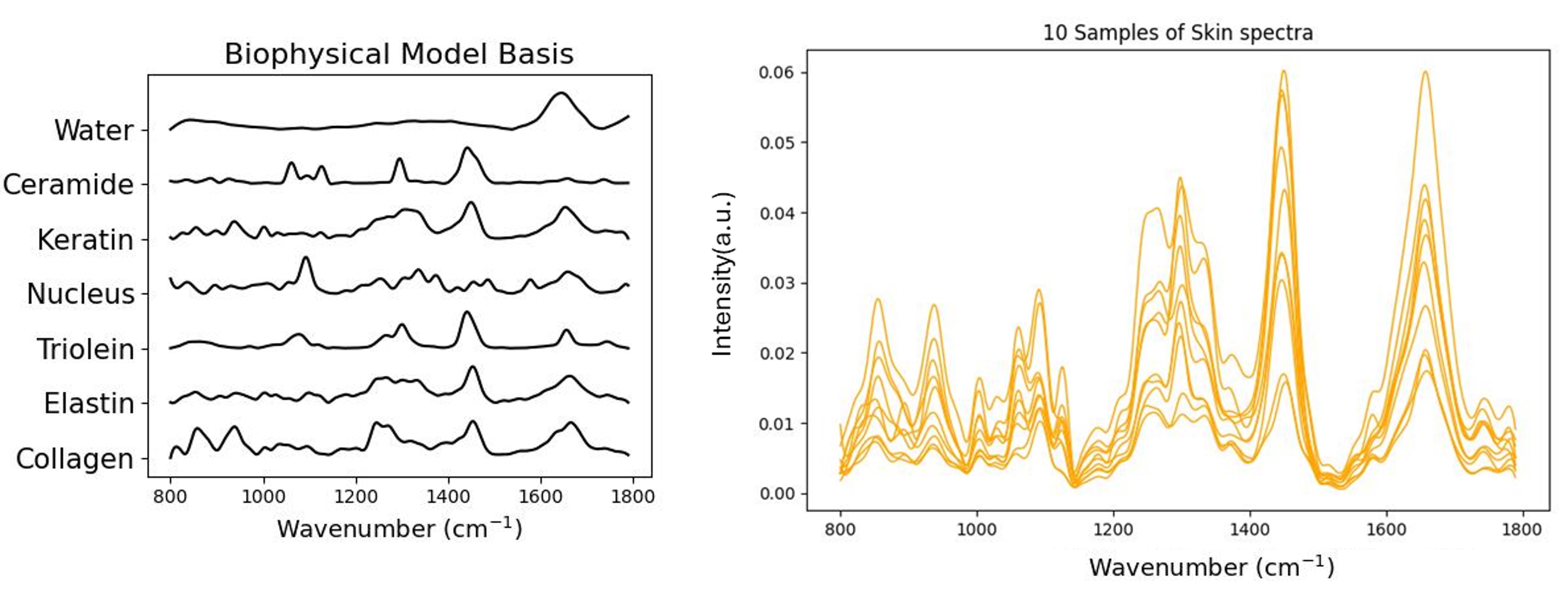}
\caption{
\textbf{Biophysical model basis and examples of simulated human skin spectra.} 
(\textbf{Left}) Area-normalized Raman spectra of seven biophysical skin components. 
(\textbf{Right}) Ten examples of simulated skin spectra generated by random weighted combinations of the basis components.
}
\label{fig:skin_spectrum_example_appendix}
\end{figure}

Figure~\ref{fig:skin_spectrum_example_appendix} illustrates the foundation of our simulated skin spectra. On the left, we display the Raman spectra of seven major biophysical components of human skin: water, ceramide, keratin, nucleus, triolein, elastin, and collagen. These basis spectra were experimentally acquired and reported by Feng et al.~\cite{Xu_biophysical}, and serve as the building blocks for synthesizing realistic composite skin spectra.

On the right, we show ten examples of simulated skin spectra generated by linearly combining the basis components with random weights. This simulates biological variability across different tissue samples. The resulting spectra exhibit diverse peak structures and intensity profiles, closely resembling true Raman signals obtained from skin.

\section*{Acknowledgments}

The authors acknowledge the use of ChatGPT (OpenAI) to assist with language clarity and editorial refinement in the preparation of this manuscript.


\bibliography{report}   
\bibliographystyle{spiejour}   


\vspace{2ex}\noindent\textbf{Mengkun Chen} is a Post-doctoral fellow at the University of Texas at Austin. He received his BS in Materials Science from Nanjing University, China, MS degree in Biomedical Engineering from Duke University and his PhD degree in Biomedical Engineering from the University of Texas at Austin in 2025. His current research interests include optical imaging, image and signal processing, and Artificial Intelligence (AI) in medical image.

\vspace{1ex}
\noindent Biographies and photographs of the other authors are not available.

\listoffigures
\listoftables

\end{spacing}
\end{document}